\documentclass{article}
\usepackage{spconf,amsmath,graphicx}
\usepackage[english]{babel}
\usepackage{graphicx}
\usepackage{subfigure}
\usepackage{amsmath,amsfonts,amssymb}
\usepackage{algorithm}
\usepackage{algpseudocode}
\usepackage{epstopdf}
\usepackage{color}
\usepackage[normalem]{ulem}
\usepackage{amsthm}
\usepackage{amsbsy}
\usepackage{varioref}
\usepackage{cases}
\usepackage{verbatim}
\usepackage{cite}
\usepackage{breqn}
\usepackage{dblfloatfix}

\title{X-ray Image Separation via Coupled Dictionary Learning}
\name{Nikos Deligiannis$^{a,b}$, Jo\~ao~F.~C.~Mota$^{c}$, Bruno~Cornelis$^{a,b}$, Miguel~R.~D.~Rodrigues$^{c}$,
        Ingrid~Daubechies$^{d}$\thanks{The work is supported by the VUB  research programme M3D2, the EPSRC grant EP/K033166/1, and the VUB-UGent-Duke-UCL Int. Joint Group on Big Data. We acknowledge valuable feedback from Jonathon Chambers.}}
\address{$^{a}$Department of Electronics and Informatics, Vrije Universiteit Brussel, Brussels 1050, Belgium.\\
$^{b}$iMinds vzw, Ghent, Belgium.
\\
$^{c}$Department of Electronic and Electrical Engineering, University College London, UK.
\\
$^{d}$Department of Mathematics, Duke University, Durham, NC 27708 USA.}
\begin{document}

%
 \maketitle%
\begin{abstract}
In support of art investigation, we propose a new source separation  method that unmixes a single  X-ray scan acquired from double-sided paintings. Unlike prior source separation methods, which are based on  statistical or structural incoherence of the sources, we use visual images taken from
the front- and  back-side of the panel to drive the separation process. The coupling of the two imaging modalities is achieved via a new multi-scale dictionary learning method. Experimental results demonstrate that our method succeeds in the discrimination of the sources, while state-of-the-art methods fail to do so.  
\end{abstract}
\section{Introduction}
\label{sec:intro}

The analysis and  enhancement of high-resolution digital acquisitions
of paintings is becoming a popular field of research~\cite{Maaten2015,Noord2015}. Prior  work includes the removal of canvas artifacts in high-resolution photographs \cite{cornelis2012digital}, the removal of cradling in X-ray images of paintings on panel \cite{icip14}, as well as the detection and digital removal of cracks ~\cite{cornelis2012crack}.

In this work, we propose a novel framework to separate X-ray images taken from double-sided  paintings. A famous piece of art that contains panels  painted on both sides is the \textit{Ghent Altarpiece} (1432) created by Jan and Hubert
van Eyck. In preparation of its restoration, the masterpiece was digitized by means of
various modalities: visual photography, infrared photography and reflectography, and X-radiography \cite{pizurica2015digital}.
The latter  is a powerful tool for art  investigation, as it reveals information about the structural condition of the painting. However,  X-ray scans of double-sided paintings are very  cluttered, thus making their reading by art experts difficult. The reason is that these images contain information from both sides of the painting as well as its support (wood structure or canvas).  


Prior work on umixing signals focuses mostly on the blind source separation
(BSS) problem, where the task  is to retrieve the different signal sources from one or more linear  mixtures. Independent component analysis (ICA) \cite{hyvarinen2004independent}---where the sources are assumed to be statistically independent---and nonnegative matrix factorization---where the sources are considered or transformed into a nonnegative representation
 \cite{smaragdis2014static}---are representative methods to solve the BSS problem.    
Alternative solutions adhere to  a Bayesian formulation, via, for example,
Markov random fields \cite{kayabol2009bayesian}. Sparsity is another source prior, heavily exploited in BSS problems \cite{bobin2007sparsity,zibulevsky2001blind}, with morphological component
analysis (MCA) being a state-of-the-art method. The  assumption in MCA is that each source has a different
morphology; namely, it has a sparse representation over a   set of
bases, alias, dictionaries, while being non-sparse  over  other dictionaries. 
The dictionaries can be pre-defined, for instance,  the wavelet or the discrete cosine transform (DCT), or  learned from a set of training signals. Seminal dictionary learning works include the method of optimal directions
(MOD) \cite{engan1999method} and the K-SVD algorithm \cite{aharon2006img},
both utilizing the orthogonal matching pursuit (OMP) \cite{tropp2007signal}
method to perform  sparse signal decomposition. Recently, MCA has been
combined with K-SVD, thus enabling dictionaries to be learned
 while separating \cite{abolghasemi2012blind}.

The  assumptions in previous source separation methods are not
fitting our  problem as both  sources have similar morphological
and statistical traits. In this work, we propose a novel method
to perform  separation of X-ray images of paintings by using images of
another modality as side information. Our approach consists of two steps: 1) learning multi-scale dictionaries from photographs and X-rays of single-sided panels (in which the X-rays are not mixed), and 2) separating the given mixed
X-ray from a double-sided panel, using those dictionaries and the  photographs from each side. Previous work has used coupled dictionary learning to address problems in audio-visual analysis \cite{monaci2007learning}, super-resolution \cite{yang2012coupled}, photo-sketch synthesis\cite{wang2012semi}, and   human pose estimation  \cite{jia2010factorized}.  Besides the application domain, our method differs from prior work in the way we model the correlation between the sources. 
Experimental evidence proves that our method is superior compared to the state-of-the-art MCA technique, configured  either with fixed or trained dictionaries.
\section{ Image Separation with Side Information} 
\label{sec:SourceSepSideInfor}
We start by describing MCA, as the state-of-the-art sparsity based source  separation  method, and afterwards we introduce the proposed method which, unlike the former, makes use of side information. First, let us denote by $x_1\in\mathbb{R}^{n\times
1}$ and $x_2\in\mathbb{R}^{n\times
1}$ two vectorized X-ray image patches that we wish to separate from a given X-ray scan patch $m=x_1+x_2$.
\\
\textbf{Morphological Component Analysis.} 
Assume that each $x_{i}$ admits a sparse decomposition in a different overcomplete dictionary $\Lambda_{i}\in\mathbb{R}^{n\times
d_{i}}$, $(n\ll\ d_i)$; namely, each component can be expressed as $x_i
= \Lambda_i z_i$, where $z_i\in\mathbb{R}^{d_{i}\times1}$ is a sparse vector
comprising a few non-zero coefficients: $\|z_i\|_0=s_i\ll
d_i$, with  $\|\cdot\|_0$ denoting the $\ell_0$ pseudo-norm. MCA~\cite{bobin2007sparsity,zibulevsky2001blind}
 decomposes the  mixture  by approximately solving the following  optimization problem:         \begin{align}\label{Eq:mcaProb}
                \begin{array}{ll}
                        \underset{z_{1}, z_{2}}{\text{minimize}}
& \|z_{1}\|_0 + \|z_{2}\|_0 \\
                        \text{subject to} & m = \Lambda_{1}z_{1} + \Lambda_{2}z_{2}\\
                \end{array}             
        \end{align}
A typical approximation consists of replacing the $\ell_0$ pseudo-norm with the $\ell_1$-norm.
\\
\textbf{Source Separation with Side Information.}
The use of side information has proven beneficial in various inverse problems \cite{vaswani2010modified,mota2014comp,mota2014compressed}. Adhering to this logic, we show how side information can be helpful in separating mixtures, where the sources have similar characteristics.  In our particular problem, we consider side information signals  $y_1$ and $y_2$ formed by the co-located visual image
 patches of the
front and the back of the painting. Both the X-ray and  side information signals admit a sparse decomposition
in given dictionaries, namely,
\begin{align}\label{eq:visual_model}
y_{1} &=\Psi^{c}z_{1c}  \nonumber\\
y_{2} &=\Psi^{c}z_{2c}  \ ,
\end{align}
and
\begin{align}\label{eq:xray_model}
x^{ray}_{1} &=\Phi^{c}z_{1c} + \Phi v \nonumber\\
x^{ray}_{2} &=\Phi^{c}z_{2c} + \Phi v_,
\end{align}
where $z_{ic}\in \mathbb{R}^{\gamma\times1}$, with $\|z_{ic}\|_0=s_{z}\ll \gamma$, denotes the sparse component  that is common
to the visual and X-ray images with respect to dictionaries
$\Psi^{c},\Phi^{c}\in\mathbb{R}^{n\times{\gamma}}$. Moreover, $v\in \mathbb{R}^{d\times{1}}$, with $\|v\|_0=s_v\ll d$, denotes the sparse innovation component of the X-ray image, obtained with
respect to dictionary $\Phi\in\mathbb{R}^{n\times{d}}$. The common components express the  structure  underlying both the X-ray and 
natural images, while the innovation component captures X-ray specific parts
of the signal (e.g., traces
of the wooden panel).
The separation 
problem is now formulated as the following   problem: 
        \begin{align}\label{Eq:Prob}
                \begin{array}{ll}
                        \underset{z_{c1}, z_{c2},v}{\text{minimize}}
& \|z_{c1}\|_0 + \|z_{c2}\|_0 +\|v\|_0  \\
                        \text{subject to} & m = \Phi^{c}z_{1c} + \Phi^{c}z_{2c}
+ 2\Phi v \\
& y_{1} =\Psi^{c}z_{1c} \\
& y_{2} =\Psi^{c}z_{2c}
                \end{array}             
        \end{align}
The relaxed version of Problem \eqref{Eq:Prob}  boils down to Basis Pursuit, which is solved by convex optimization tools, e.g., \cite{BergFriedlander2008}.
\begin{algorithm}[t]  
    \caption{modified Orthogonal Matching Pursuit }
    \label{Alg:modifiedOMP}
    \begin{algorithmic}[1]
    \small
     \Statex
     \algrenewcommand\algorithmicrequire{\textbf{Initialization}}
     \Require           
\State Initialize  residual: $r_0=b$.
\State Total sparsity of vector $w$: $s_w=s_z+s_v$.
\State Counters for the sparsity of $z$ and $v$: $\ell_z=0$, $\ell_v=0$.
\State Set of non-zero elements of $w$: $\Omega=\emptyset$.                  
                \Statex                                                 
                \algrenewcommand\algorithmicrequire{\textbf{Algorithm}}
                \Require                                    
\For{$i = 1,2,\dots, s_w$}                          
\State Sort the indices $\zeta=\{1,2,\dots,\gamma+d\}$, corresponding to
the $\theta_{\zeta}$ columns of $\Theta$,  such that $|\langle r_{i-1}, \theta_{\zeta} \rangle|$ are in descending order. Put the ordered indices in the
vector
 $q_{i}$.
\State Set $\mathcal{G}=\emptyset$ and auxiliary iterator $\texttt{it}=0$.
           \While{$\mathcal{G}=\emptyset$}
                \State{\texttt{it}=\texttt{it}+1.}           
                \State Find index that corresponds
to value of \texttt{it}: {$\kappa = q_i[\texttt{it}]$.} 
                \If  {$\kappa\in\mathcal{I}$ \text{AND} $\ell_z<s_z$}
                 \State {Set $\mathcal{G}=\kappa$ and increase: $\ell_z=\ell_z
+1$.}
                 \Else
                 \If{$\kappa\in\mathcal{J}$ \text{AND} $\ell_v<s_v$}
                \State {Set $\mathcal{G}=\kappa$ and increase : $\ell_v=\ell_v
+1$.}
                 \EndIf
                 \EndIf
                 \EndWhile
                 \State Update the set of non-zero elements of $w$, i.e.,
$\Omega_{i} = \Omega_{i-1}\cup\{\kappa\}$, and the matrix of chosen atoms:
$\Theta_i=[\Theta_{i-1}\quad \theta_{\kappa}]$.
                 \State Solve: $w_i=\arg\min_w\|b-\Theta_i
w\|_2$.
                 \State Calculate the new residual: $r_i=b-\Theta_i w_i$.

                 \EndFor
    \end{algorithmic}
  \end{algorithm}

\section{Coupled Dictionary Learning Algorithm}
\label{sec:coupledDicLearn}
We train coupled dictionaries, $\Psi^c$, $\Phi^c$, $\Phi$, by using image patches  sampled from registered
visual and X-ray images of single-sided  panels, which 
do not suffer from superposition phenomena.
Let $Y,X \in \mathbb{R}^{n \times
t}$ represent a set of~$t$ co-located vectorized
visual and X-ray patches, each
containing $\sqrt{n}\times\sqrt{n}$
 pixels. We assume that the columns of $X$~and $Y$~can be decomposed as in~\eqref{eq:xray_model}, and we collect their common components into the columns of the matrix $Z \in \mathbb{R}^{\gamma \times t}$ and their innovation components into the columns of $V\in\mathbb{R}^{d \times{t}}$. 
We formulate the coupled dictionary learning problem as
        \begin{equation}\label{Eq:Problem}
                \begin{array}[t]{cl}
                        \underset{\begin{subarray}\\ \Psi^c , Z \\ \Phi^c,\Phi,V\end{subarray}}{\text{minimize}}
                        &
                                \frac{1}{2}\big\|Y - \Psi^c Z \big\|_F^2
                                +
                                \frac{1}{2}\big\|X - \Phi^c Z - \Phi V\big\|_F^2,
                        \\
                        \text{subject to} &
                                 \big\|z_\tau\big\|_0 \leq s_{z},  
                          
                                \vspace{0.2cm}
                                \\&
                                \big\|v_\tau\big\|_0\leq s_{v}, \quad \forall
\tau=1,2,\dots t,
                \end{array}
        \end{equation}
        where  $z_\tau$, $v_\tau$ are sparse vector-columns of matrix $Z$ and $V$, $\tau$ runs over the columns of $Z$ and $V$, and $s_{z}$, $s_{v}$ are thresholds on the sparsity level.  Given initial estimates for the dictionaries\footnote{We use the overcomplete DCT to initialize our dictionaries.}, Problem~\eqref{Eq:Problem} is solved by iterating between a sparse-coding step, where the dictionaries are fixed, and a dictionary update step, in which the coefficients are fixed, as in
 \cite{aharon2006img,engan1999method}. 

        Given fixed dictionaries, the sparse coding problem decomposes into~$t$ problems that can be solved in parallel:
       \begin{equation}
(z_\tau^{k+1}, v_\tau^{k+1}) =
                \begin{array}[t]{cl}
                        \underset{\quad\; z_\tau,v_\tau}{\text{arg}\;\text{min}} &
\frac{1}{2}
                        \Bigg\|
                                \begin{bmatrix}y_{\tau} \\ x_{\tau}\end{bmatrix}

                                -
                                \begin{bmatrix}
                                        {\Psi^c}^k &  0 \\
                                        {\Phi^c}^k & \Phi^k
                                \end{bmatrix}
                                \begin{bmatrix}
                                        z_\tau  \\ v_\tau
                                \end{bmatrix} \vspace{0.2cm}
                        \Bigg\|_F^2,
                        \\
                        \text{subject to} & \big\|z_\tau\big\|_0 \leq s_{z},     
                                \vspace{0.2cm}
                                \\&
                                \big\|v_\tau\big\|_0\leq s_{v}, \quad \forall
\tau=1,2,\dots t,
                \end{array}             
                \label{Eq:SolverWeights1}
        \end{equation} 
where we used~$x_\tau$, $y_\tau$, $z_\tau$, and~$v_\tau$ to represent column~$\tau$
of~$X$, $Y$, $Z$, and~$V$, respectively, and $k$ counts the iterations. To address each of the $t$ sub-problems
in~\eqref{Eq:SolverWeights1}, we propose a greedy algorithm that constitutes
a modification of the   OMP method [see Algorithm \ref{Alg:modifiedOMP}]. 
Our method adapts OMP~\cite{tropp2007signal} to solve:
\begin{align}
\begin{array}{cc}
\underset{w}{\text{minimize}} & \|b - \Theta w\|_2^2 \\
\text{subject to} & \|(w(\mathcal{I})\|_0\leq s_z, \\
 & \|w(\mathcal{J})\|_0\leq s_v,
\label{Eq:ompProb}
\end{array}
\end{align}
where $w(\mathcal{I})$ [resp., $w(\mathcal{J})$] denotes the components
of  vector $w\in \mathbb{R}^{(\gamma+d)\times1}$ indexed by the index set $\mathcal{I}$
(resp., $\mathcal{J}$), with $\mathcal{I}\cup\mathcal{J}=\{1,2,\dots,\gamma+d\}$,
$\mathcal{I}\cap\mathcal{J}=\{\O\}$. Each sub-problem
in \eqref{Eq:SolverWeights1} translates to \eqref{Eq:ompProb}  
by replacing: $b=\begin{bmatrix}y_\tau \\ x_\tau\end{bmatrix}$,
$\Theta=\begin{bmatrix}
        {\Psi^c}^k & 0 \\
        {\Phi^c}^k & \Phi^k
        \end{bmatrix}$,
and $w=\begin{bmatrix} z_\tau \\ v_\tau\end{bmatrix}$.

\begin{figure}[t]
 \centering
 \subfigure[]{
    \label{fig:XrrSIvisuals:a}
    \includegraphics[width=4cm]{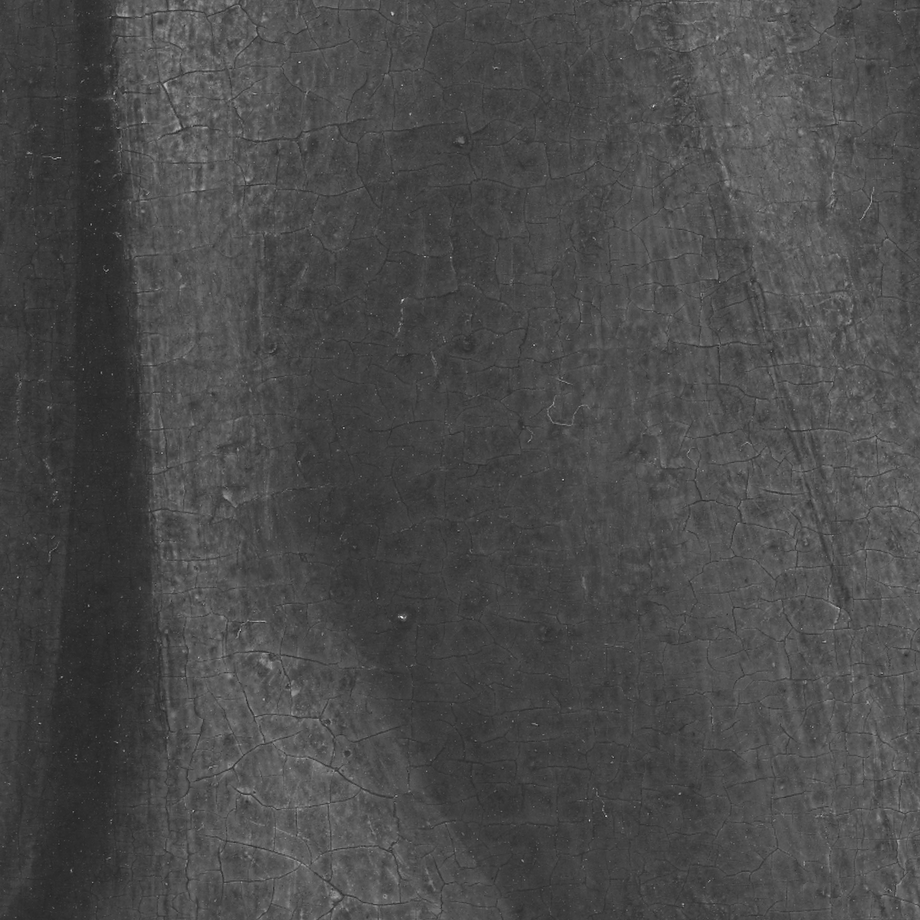}}
 \subfigure[]{
    \label{fig:XrrSIvisuals:b}
    \includegraphics[width=4cm]{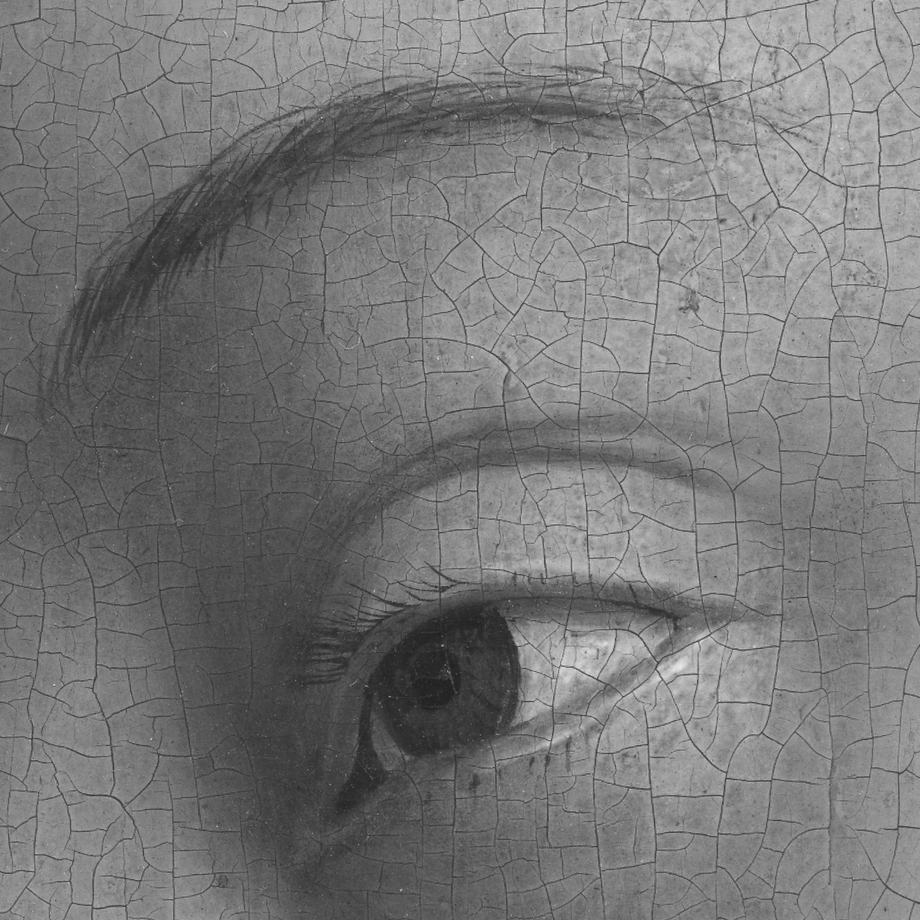}}
  \subfigure[]{
    \label{fig:XrrSIvisuals:a}
    \includegraphics[width=4cm]{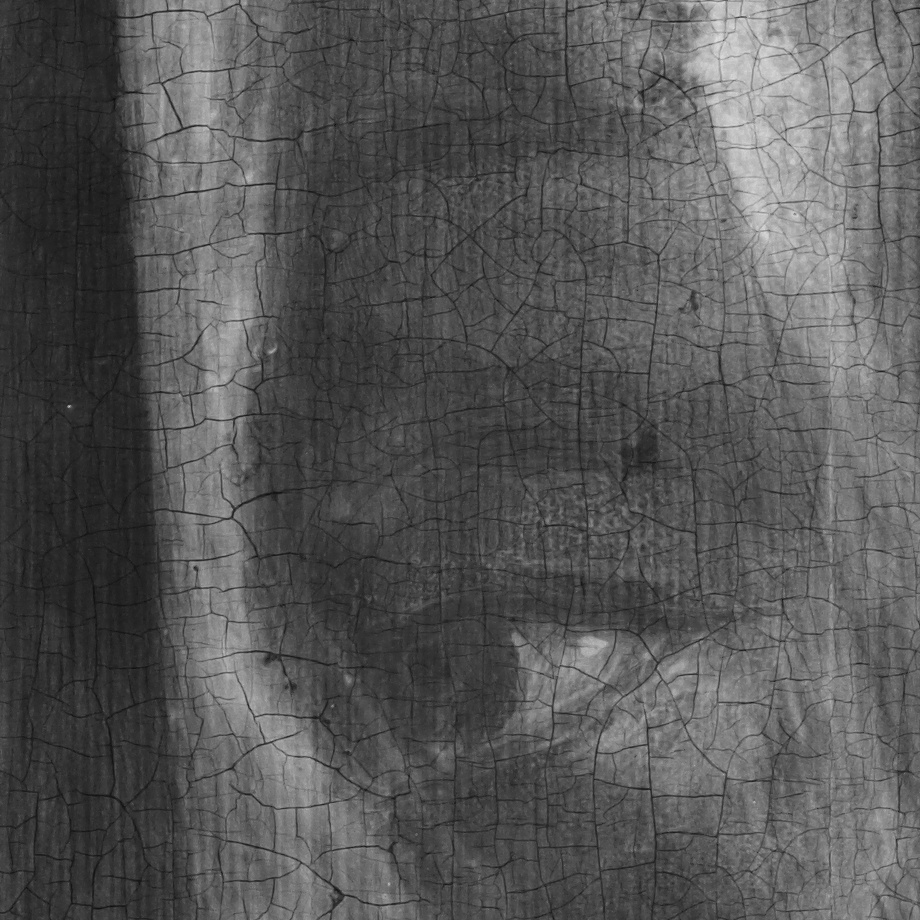}}
  \caption{Image set cropped from a double-sided panel of the altarpiece,  on which we assess the proposed method; (a) photograph of side 1, (b) photograph of side 2; (c) corresponding  X-ray image. The resolution is $1024\times{1024}$ pixels.}
\label{fig:XrrSIvisuals}
\end{figure}   
     
Given fixed sparse coefficients, the dictionary update  problem decouples into two (independent)
problems, that is,
                $$
                        {\Psi^c}^{k+1}=\arg\min_{\Psi^c}
                        \frac{1}{2}\Big\|Y - \Psi^c\cdot Z^{k+1}\Big\|_F^2
$$and$$
                        \overline{\Phi}^{k+1}=\arg\min_{\overline{\Phi}}
                        \frac{1}{2}\Big\|X - \overline{\Phi}\cdot\overline{V}^{k+1}\Big\|_F^2,
                $$
where $\overline{\Phi} = \begin{bmatrix}\Phi^c & \Phi\end{bmatrix}$ and $\overline{V}^{k+1}
= \begin{bmatrix}Z^{k+1} \\ V^{k+1}\end{bmatrix}$.
Each of these problems has a closed-form solution. 
\begin{figure}[t!]
 \centering
  \subfigure{
    \label{fig:visualexample:a}
    \includegraphics[width=4cm]{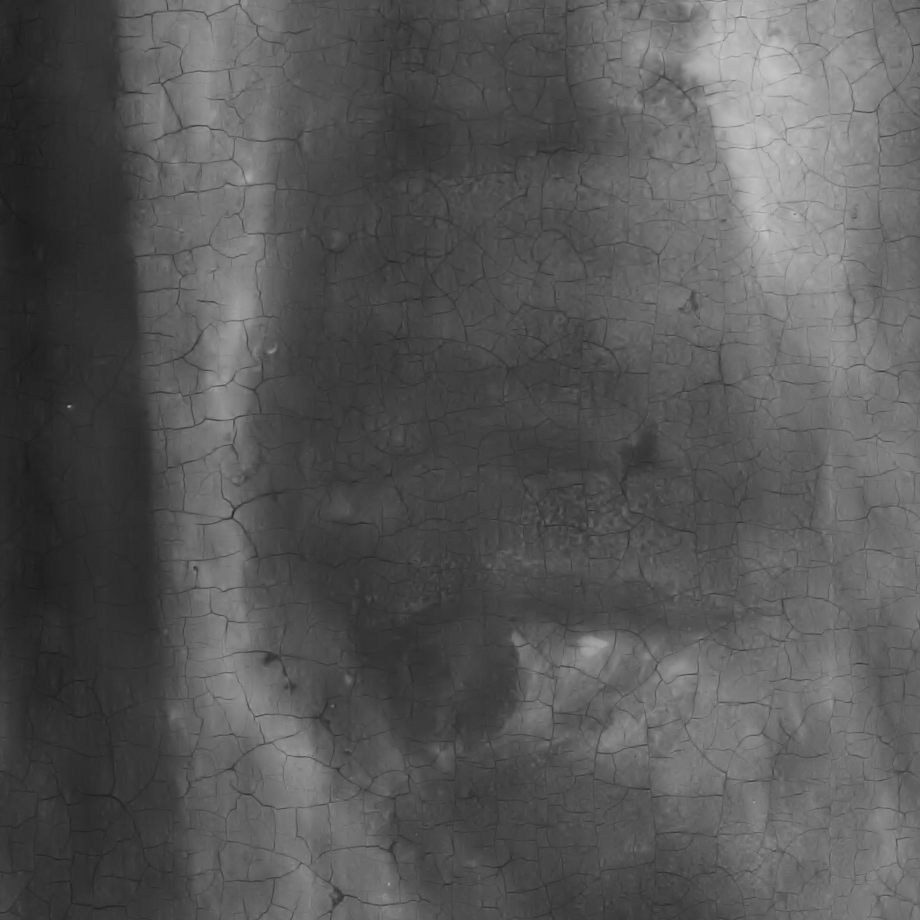}}
  \subfigure{
    \label{fig:visualexample:b}
    \includegraphics[width=4cm]{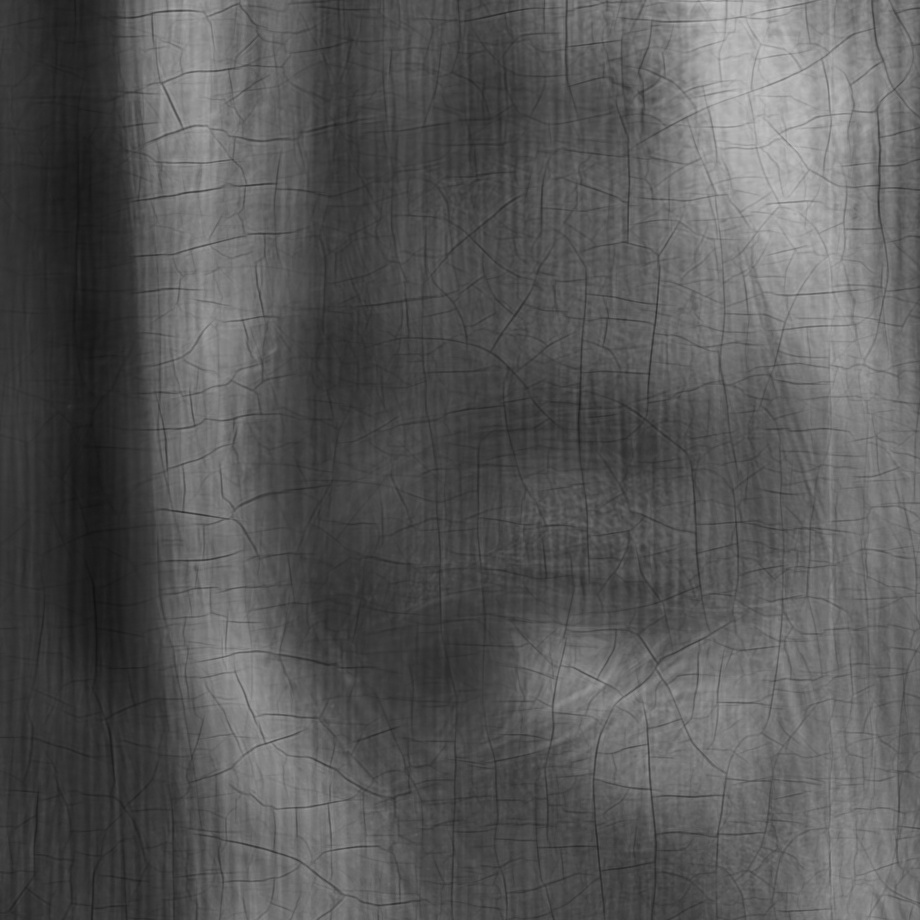}}
  \\\subfigure{
    \label{fig:visualexample:c}
    \includegraphics[width=4cm]{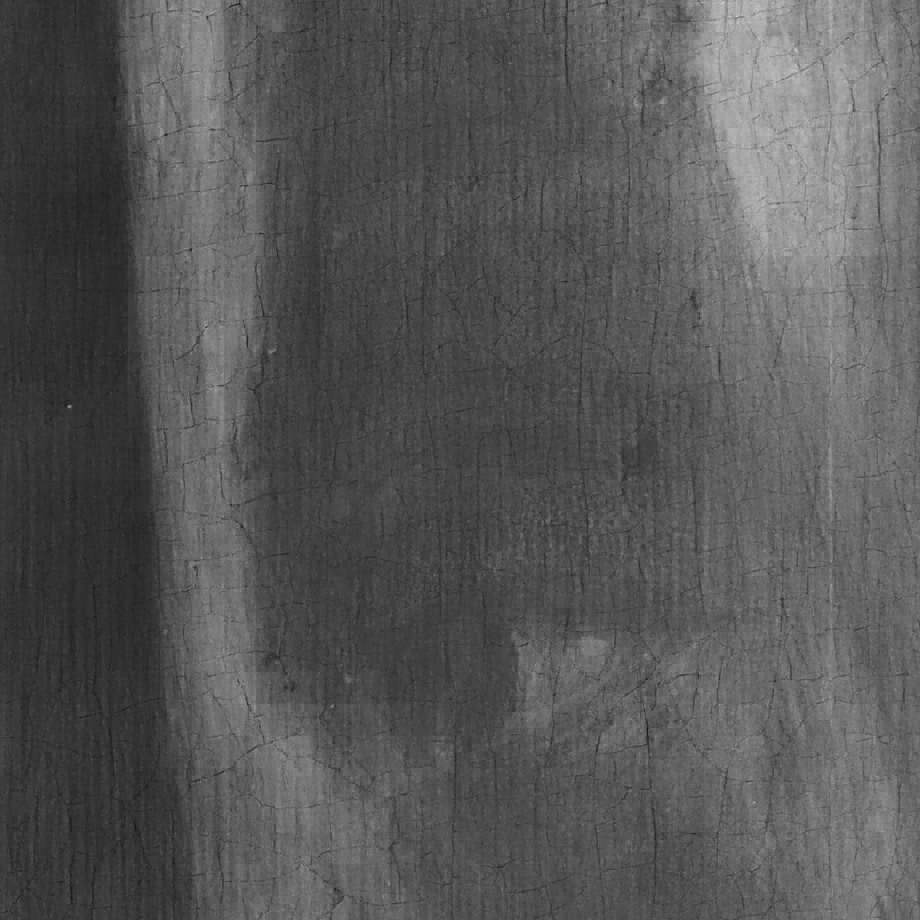}}
  \subfigure{
    \label{fig:visualexample:d}
    \includegraphics[width=4cm]{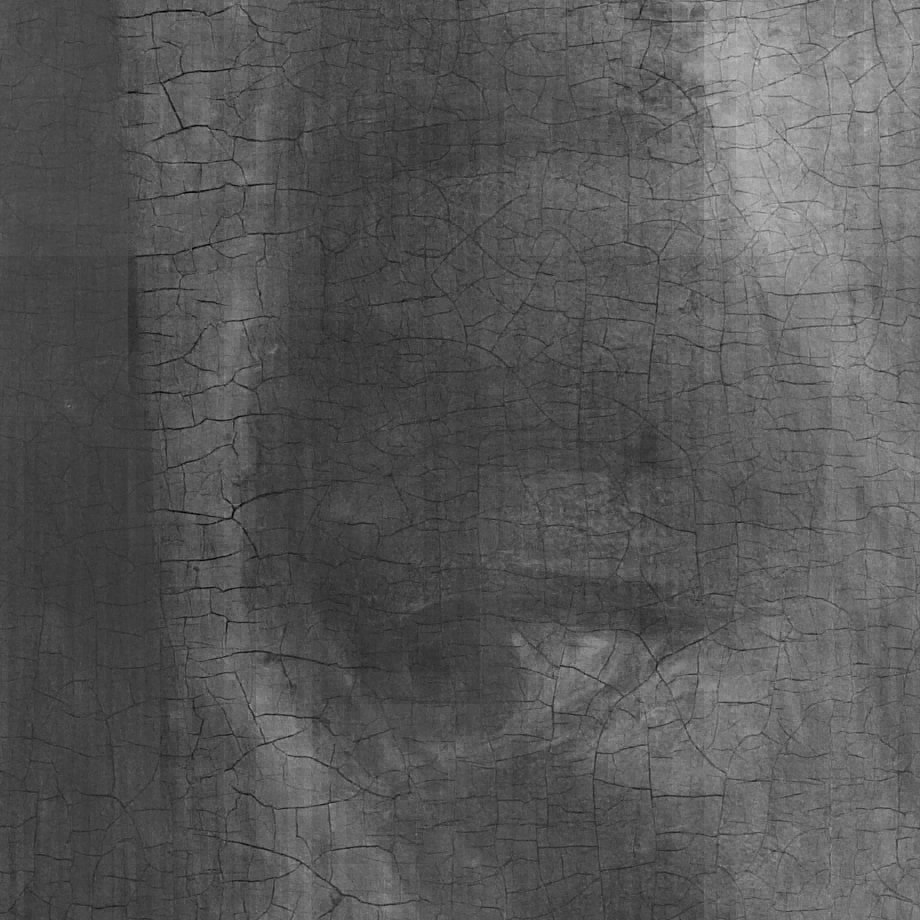}}
  \\\subfigure{
    \label{fig:visualexample:c}
    \includegraphics[width=4cm]{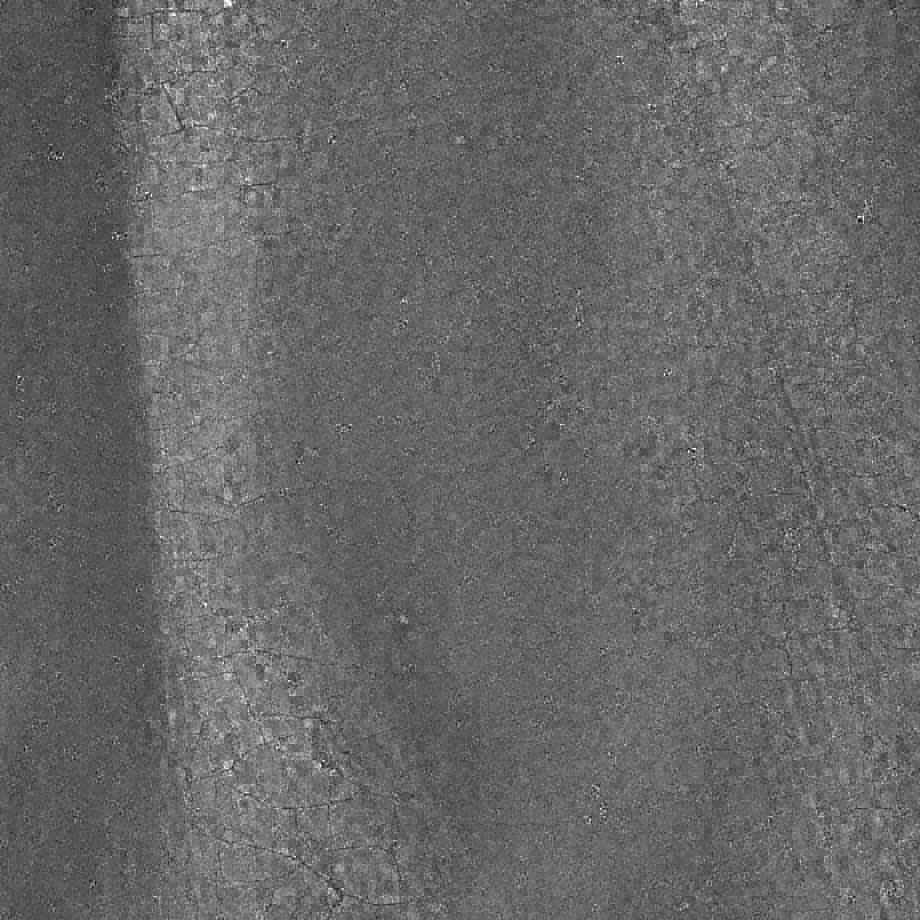}}
  \subfigure{
    \label{fig:visualexample:d}
    \includegraphics[width=4cm]{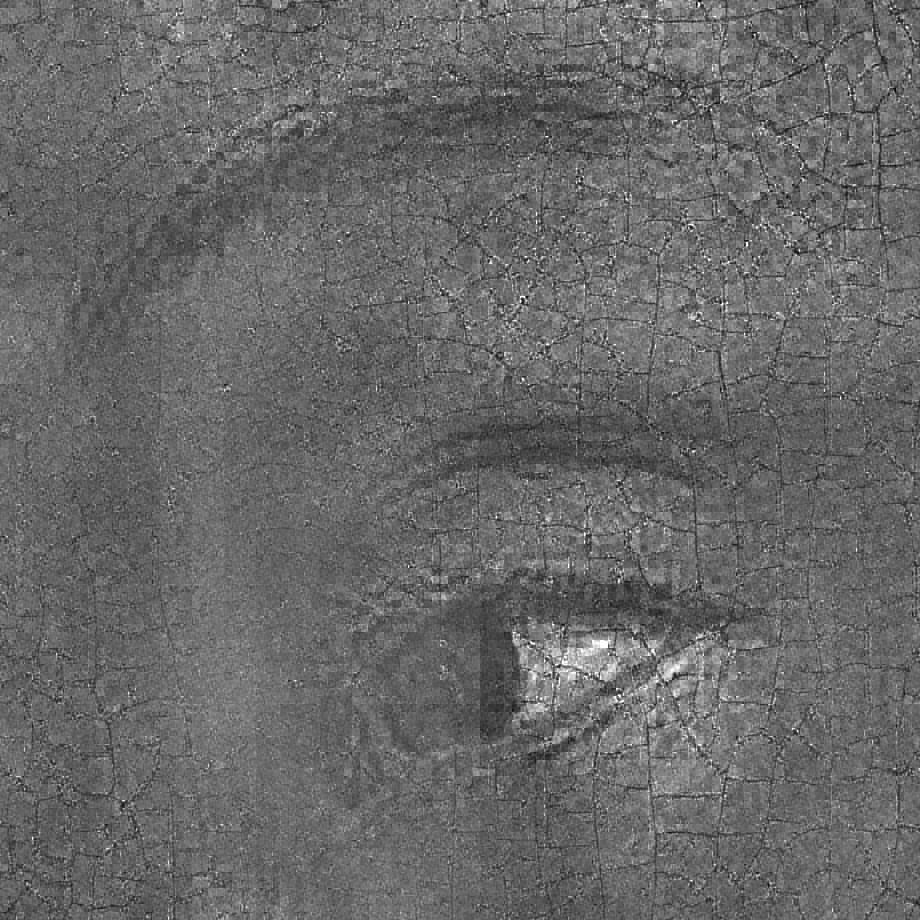}}
  \caption{Visual evaluation of the proposed multi-scale
method in the separation of the X-ray in Fig. \ref{fig:XrrSIvisuals}(c);
(left) separated side 1, (right) separated side 2. The competing methods
are: (1st row) MCA\ with fixed dictionaries \cite{icip14}, (2nd row) multi-scale
MCA with K-SVD, (3rd row) Proposed.}
\label{fig:visualexample2}
\end{figure}

\section{X-ray Image Separation Method}
\label{sec:MultScAppr}
Because of  complexity, dictionaries are learned for small image patches, usually with dimensions of $8\times8$ pixels; namely, we adhere to a local sparsity prior. However, due to the high-resolution of the images, patches of that size cannot fully capture  large structures.   Hence, we propose a multi-scale image separation approach that is based on a pyramid decomposition of the images. Our multi-scale strategy is as follows: The  images  at scale $l=\{1, 2,\dots,L\}$---where we use the notation
$M_l, Y_{1,l}, Y_{2,l}$, to refer to the mixed X-ray and the
two visuals, respectively---are divided into overlapping patches  $m_l^{\boldsymbol{u}_l},
y_{1,l}^{\boldsymbol{u}_l}, \:\text{and}\: y_{2,l}^{\boldsymbol{u}_l}$, each
of size $\sqrt{n_l}\times\sqrt{n_l}$ pixels. Each patch has top-left coordinates
$$
\boldsymbol{u}_l = (\epsilon_l\cdot u_{1,l}, \epsilon_l\cdot u_{2,l}), \:
0\leq u_{1,l}<\left\lfloor
\frac{H_l}{\epsilon_l} \right\rfloor, \: 0\leq u_{2,l}<\left\lfloor
\frac{W_l}{\epsilon_l} \right\rfloor,
$$
where $\epsilon_l\in\mathbb{Z}_+, \: \sqrt{n_l}\leq \epsilon_l<\sqrt{n_l}$
is the overlap step-size, and  $H_l, W_l$ are the height and width  of the
image decomposition at
scale $l$. The DC value is extracted from  each patch, thereby constructing
the high frequency  band of the image  at scale $l$. The aggregated
DC values comprise the low-pass component of the image, the resolution of
which  is $\left\lfloor\frac{H_l}{\epsilon_l} \right\rfloor\times\left\lfloor
\frac{W_l}{\epsilon_l} \right\rfloor$ pixels. The low-pass component is then
decomposed further at the subsequent scale ($l+1$). The texture of the mixed X-ray image  at scale $l$ is separated patch-per-patch
by solving Problem  \eqref{Eq:Prob}. 
The texture  of
each separated  patch is then reconstructed as $x_{1,l}^{\boldsymbol{u}_l} = \Phi^{c}_lz_{1c,l}^{\boldsymbol{u}_l}
$ and $x_{2,l}^{\boldsymbol{u}_l} = \Phi_l^{c}z_{2c, l}^{\boldsymbol{u}_l}$.
Namely, we omit the innovation component $v$ [see \eqref{eq:xray_model}] during reconstruction, as this is common to the two X-rays\footnote{Experimental observation revealed that including the innovation component leads to poorer visual quality of the separation. }. 
The separated X-ray images are finally reconstructed by following the reverse
operation: Descending the pyramid, the separated component
at the coarser level is up-sampled and added to the separated component of
the finer scales.

As a final note, the dictionary learning process is applied per scale, yielding
a triple of coupled dictionaries $(\Psi^c_l,\Phi^c_l,\Phi_l)$ per scale $l$.
Due to  lack of training data in the coarser scales, dictionaries
are typically learned on the finer scales and then re-used
in the coarsest scale.

\section{Experiments}
\label{Sec:ExpSection}
We assess our method on different crops, with dimensions of $1024\times1024$, taken from the digital acquisitions~\cite{pizurica2015digital} of one  double-sided panel of the \textit{Ghent Altarpiece} (1432). An example  X-ray image we aim to separate and the  two corresponding
visual images  from each side of the panel are depicted in Fig. \ref{fig:XrrSIvisuals}.
We apply the  multi-scale framework,
where we use $L=3$ scales with parameters  $\sqrt{n_l} =
8$, $\epsilon_1 = 4$, $\epsilon_2=4$ and $\epsilon_3= 7$.
 Dictionary triplets $(\Psi^c_\ell,\Phi^c_\ell,\Phi_\ell)$, each with dimension of $64\times{256}$,  are trained for the first two layers and the dictionaries
of the second layer are extrapolated to the third. We use $t=46000$ patches from digital acquisitions of the  single-sided panels of the altarpiece and set $s_{z}=10$ and $s_{v}=8$.

To demonstrate the
benefit of using side information, we compare our method against two configurations of MCA~\cite{bobin2007sparsity,zibulevsky2001blind}. In the first one we use the discrete
wavelet and curvelet transforms on blocks of $512\times 512$
pixels \cite{icip14}; the low-frequency
content is divided between the two components. In the second configuration we use K-SVD to train two dictionaries: one on X-ray images
depicting  cloth and the other on images depicting faces---content also found in the  X-ray mixtures. The K-SVD\ 
method is extended with our multi-scale strategy and the same parameters are used. As no ground truth data is available, we first resort to visual comparisons. The results, depicted in~Fig.~\ref{fig:visualexample2},  clearly show that MCA with fixed dictionaries can only separate based
on morphological properties; for example, the wood grain of the panel is
captured  entirely by curvelets and not by the wavelets. It is, however,
unfitted to separate painted content. MCA with K-SVD  dictionaries
is also unable to separate the X-ray content as the dictionaries are not sufficiently discriminative. 
 The results using our method show the benefit of incorporating side information in the separation problem.
Towards a more objective comparison, we measure the structural similarity (SSIM) \cite{wang2004image} index between the two separated components, where  low SSIM\ values would indicate less similarity; hence good separation.   The results on two additional X-ray scans from the same painting, reported in Table \ref{tab:SimulMixtureTable},  confirm the better separation performance of our method, as advocated by the lowest SSIM values.

\begin{table}[t]
\caption{Similarity scores (obtained with the SSIM~\cite{wang2004image} metric) between the
separated components. 
} 
\label{tab:SimulMixtureTable}
\centering 
\tabcolsep=0.05cm
\footnotesize
\begin{tabular}{c|c|c|c} 
\hline\hline
          & MCA fixed & MCA trained & Proposed\\[0.1ex]
\hline 
X-ray mixture 1 & 0.9249 & 0.7385 & 0.1681 \\
\hline
X-ray mixture 2 & 0.9603 & 0.8341 & 0.6034 \\
\hline\hline 
\end{tabular}
\end{table}

\section{Conclusion}
\label{sec:conclusion}
We have proposed a novel sparsity-based regularization method for source separation guided by side information. Our method is based on a new multi-scale   algorithm that learns dictionaries coupling multi-modal data.  We apply the proposed method to separate X-ray images of paintings with content on both sides of their panel, where photographs of each side are used as side information. Experiments with real data from digital acquisitions of the \textit{Ghent Altarpiece} (1432), prove the superiority of our method compared to the state-of-the-art MCA technique~\cite{bobin2007sparsity,zibulevsky2001blind,abolghasemi2012blind}.

\bibliographystyle{IEEEbib}
\bibliography{bib_paper}

\end{document}